# Analysis of Hybrid Soft and Hard Computing Techniques for Forex Monitoring Systems

Ajith Abraham
School of Business Systems, Monash University, Clayton, Victoria 3800, Australia. Email: ajith.abraham@ieee.org

**Abstract:** In a universe with a single currency, there would be no foreign exchange market, no foreign exchange rates, and no foreign exchange. Over the past twenty-five years, the way the market has performed those tasks has changed enormously. The need for intelligent monitoring systems has become a necessity to keep track of the complex forex market. The vast currency market is a foreign concept to the average individual. However, once it is broken down into simple terms, the average individual can begin to understand the foreign exchange market and use it as a financial instrument for future investing. In this paper, we attempt to compare the performance of hybrid soft computing and hard computing techniques to predict the average monthly forex rates one month ahead. The soft computing models considered are a neural network trained by the scaled conjugate gradient algorithm and a neuro-fuzzy model implementing a Takagi-Sugeno fuzzy inference system. We also considered Multivariate Adaptive Regression Splines (MARS), Classification and Regression Trees (CART) and a hybrid CART-MARS technique. We considered the exchange rates of Australian dollar with respect to US dollar, Singapore dollar, New Zealand dollar, Japanese yen and United Kingdom pounds. The models were trained using 70% of the data and remaining was used for testing and validation purposes. It is observed that the proposed hybrid models could predict the forex rates more accurately than all the techniques when applied individually. Empirical results also reveal that the hybrid hard computing approach also improved some of our previous work using a neuro-fuzzy approach.

## I. INTRODUCTION

After the deregulation of the foreign currency exchange rate in early 1970s in the USA and other developed countries the global economy has undergone drastic change. The wind of this change failed to reach the Australian scenario immediately due to the government's fixed foreign currency exchange rate regulation. Starting from 1983 there has been considerable changes in the Australian forex market. Like Australia most of the developed and developing countries in the world welcome foreign investors. When foreign investors get access to invest in any country's bond equities, manufacturing industries, property market and other assets then the forex market becomes affected. This affect influences our everyday personal and corporate financial lives, and the economic and political fate of every country on the earth. The nature of the forex market is generally complex and volatile. The volatility or rate fluctuation depends on many factors. Some of the factors include financing government deficits, changing hands of equity in companies, ownership of real estate, employment opportunities, merging and ownership of large financial corporation or companies etc. The major attractions to the business of forex trading are threefold, namely, high liquidity, good leverage and low cost associated with actual trading. There are, of course, many other advantages attached with the dealing of forex once we get involved and understand in more details [1] [6].

There are many ways in which traders analyze the directions of the market. Whatever the method, it is always related to the activities of the price for some periods of time in the past. The pattern in which the prices move up and down tends to repeat itself. Thus, the prediction of future price movements can be plotted out by studying the history of past price movements. Of course there are still other theories to be followed if an accurate prediction is to be expected. These theories are associated with financial jargons such as: support & resistance levels, trend lines, double bottoms and double tops, technical indicators, etc. It is well known that the forex market has its own momentum and using traditional statistical techniques based on the previous market trends and parameters, it is very difficult to predict future exchange rates. Long-term prediction of exchange rates might help the policy makers and traders for making crucial decisions. We analyzed the average monthly foreign exchange rates for continuous 244 months starting January 1981 for exchange rates of 5 international currencies with respect to Australian dollar. In this paper, we report the comparative performance of neural network, neuro-fuzzy system, MARS [9], CART [10] and a hybrid CART-MARS approach. In section II and III, we provide some theoretical background on soft computing and the considered hard computing techniques followed by experimentation setup, training and test results in section IV. Some conclusions are also provided towards the end.

## II. SOFT COMPUTING

Soft computing was first proposed by Zadeh [11] to construct new generation computationally intelligent hybrid systems consisting of neural networks, fuzzy inference system, approximate reasoning and derivative free optimization techniques. It is well known that the intelligent systems, which can provide human like expertise such as domain knowledge, uncertain reasoning, and adaptation to a noisy and time varying environment, are important in tackling practical computing problems. In contrast with conventional AI techniques which only deal with precision, certainty and rigor the guiding principle of soft computing is to exploit the tolerance for imprecision, uncertainty, low solution cost, robustness, partial truth to achieve tractability and better rapport with reality.

**Artificial Neural Networks**

In the Conjugate Gradient Algorithm (CGA) a search is performed along conjugate directions, which produces generally faster convergence than steepest descent directions. A search is made along the conjugate gradient direction to determine the step size, which will minimize the performance function along that line. A line search is performed to determine the optimal distance to move along the current search direction. Then the next search direction is determined so that it is conjugate to previous search direction. The general procedure for determining the new search direction is to combine the new steepest descent direction with the previous search direction. An important feature of the CGA is that the minimization performed in one step is not partially undone by the next, as it is the case with gradient descent methods. An important drawback of CGA is the requirement of a line search, which is computationally expensive. Moller introduced the Scaled Conjugate Gradient Algorithm (SCGA) as a way of avoiding the complicated line search procedure of conventional CGA. According to the SCGA, the Hessian matrix is approximated by

$$E''(w_k)p_k = \frac{E'(w_k + \sigma_k p_k) - E'(w_k)}{\sigma_k} + \lambda_k p_k$$

where $E'$ and $E''$ are the first and second derivative information of global error function $E(w_k)$. The other terms $p_k$, $\sigma_k$ and $\lambda_k$ represent the weights, search direction, parameter controlling the change in weight for second derivative approximation and parameter for regulating the indefiniteness of the Hessian. In order to get a good quadratic approximation of $E$, a mechanism to raise and lower $\lambda_k$ is needed when the Hessian is positive definite. Detailed step-by-step description can be found in the literature [8].

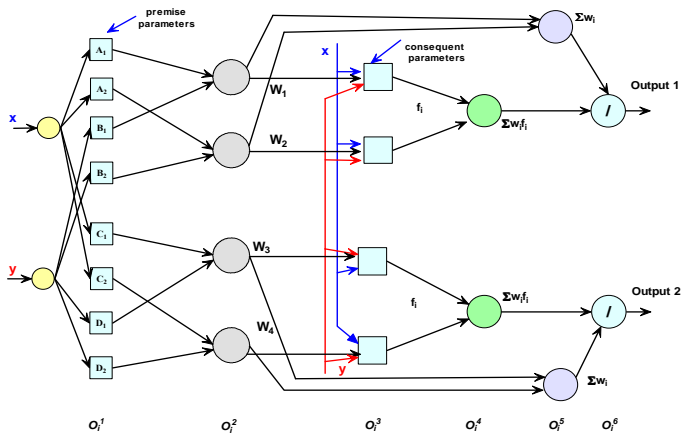

**Figure 1.** Architecture of ANFIS with multiple outputs

**Neuro-fuzzy Computing**

Neuro-Fuzzy (NF) computing is a popular framework for solving complex problems [3]. We used the Adaptive Neuro Fuzzy Inference System (ANFIS) implementing a Takagi-Sugeno type FIS. We modified the ANFIS model to accommodate the multiple outputs [1] [7]. Figure 1 depicts the 6-layered architecture of multiple output ANFIS. ANFIS uses a mixture of backpropagation to learn the premise parameters and least mean square estimation to determine the consequent parameters. A step in the learning procedure has two parts: In the first part the input patterns are propagated, and the optimal conclusion parameters are estimated by an iterative least mean square procedure, while the antecedent parameters (membership functions) are assumed to be fixed for the current cycle through the training set. In the second part the patterns are propagated again, and in this epoch, backpropagation is used to modify the antecedent parameters, while the conclusion parameters remain fixed. This procedure is then iterated [7].

### III. HARD COMPUTING

**Multivariate Adaptive Regression Splines (MARS)**

MARS is a powerful well-established adaptive regression technique, which is known for its speed and accuracy [2] [5] [9]. The MARS model is a spline regression model that uses a specific class of basis functions as predictors in place of the original data. The MARS basis function transform makes it possible to selectively blank out certain regions of a variable by making them zero, allowing MARS to focus on specific sub-regions of the data. MARS excels at finding optimal variable transformations and interactions, as well as the complex data structure that often hides in high-dimensional data. A key concept underlying the spline is the knot. A knot marks the end of one region of data and the beginning of another. Thus, the knot is where the behavior of the function changes. Between knots, the model could be global (e.g., linear regression). In a classical spline, the knots are predetermined and evenly spaced, whereas in MARS, the knots are determined by a search procedure. Only as many knots as needed are included in a MARS model. If a straight line is a good fit, there will be no interior knots. In MARS, however, there is always at least one "pseudo" knot that corresponds to the smallest observed value of the predictor. Finding the one best knot in a simple regression is a straightforward search problem: simply examine a large number of potential knots and choose the one with the best $R^2$. However, finding the best pair of knots requires far more computation, and finding the best set of knots when the actual number needed is unknown is an even more challenging task [9].

MARS finds the location and number of needed knots in a forward/backward stepwise fashion. A model, which is clearly over fit with too many knots, is generated first, then, those knots that contribute least to the overall fit are removed. Thus, the forward knot selection will include many incorrect knot locations, but these erroneous knots will eventually, be deleted from the model in the backwards pruning step (although this is not guaranteed).

In MARS, Basis Functions (BFs) are the machinery used for generalizing the search for knots. BFs are a set of functions used to represent the information contained in one or more variables. Much like principal components, BFs essentially re-express the relationship of the predictor variables with the target variable. The hockey stick BF, the core building block of the MARS model is often applied to a single variable multiple times. The hockey stick function maps variable $X^*$: max $(0, X - c)$, or max $(0, c - X)$ where $X^*$ is set to 0 for all values of $X$ up to some threshold value $c$ and $X^*$ is equal to $X$ for all values of $X$ greater than $c$. (Actually $X^*$ is equal to the amount by which $X$ exceeds threshold *c*). The second form generates a mirror image of the first.

MARS generates basis functions by searching in a stepwise manner. It starts with just a constant in the model and then begins the search for a variable-knot combination that improves the model the most (or, alternatively, worsens the model the least). The improvement is measured in part by the change in Mean Squared Error (MSE). Adding a basis function always reduces the MSE. MARS searches for a pair of hockey stick basis functions, the primary and mirror image, even though only one might be linearly independent of the other terms. This search is then repeated, with MARS searching for the best variable to add given the basis functions already in the model. The brute search process theoretically continues until every possible basis function has been added to the model. In practice, the user specifies an upper limit for the number of knots to be generated in the forward stage. The limit should be large enough to ensure that the true model can be captured. A good rule of thumb for determining the minimum number is three to four times the number of basis functions in the optimal model. This limit may have to be set by trial and error.

**Classification and Regression Trees (CART)**

Tree-based models are useful for both classification and regression problems [4]. In these problems, there is a set of classification or predictor variables ($X_i$) and a dependent variable (*Y*). The $X_i$ variables may be a mixture of nominal and / or ordinal scales (or code intervals of equal-interval scale) and *Y* a quantitative or a qualitative (i.e., nominal or categorical) variable.

The CART methodology is technically known as binary recursive partitioning [10]. The process is binary because parent nodes are always split into exactly two child nodes and recursive because the process can be repeated by treating each child node as a parent. The key elements of a CART analysis are a set of rules for:

- splitting each node in a tree;
- deciding when a tree is complete; and
- assigning each terminal node to a class outcome (or predicted value for regression)

CART is the most advanced decision-tree technology for data analysis, preprocessing and predictive modeling. CART is a robust data-analysis tool that automatically searches for important patterns and relationships and quickly uncovers hidden structure even in highly complex data. CART's binary decision trees are more sparing with data and detect more structure before further splitting is impossible or stopped. Splitting is impossible if only one case remains in a particular node or if all the cases in that node are exact copies of each other (on predictor variables). CART also allows splitting to be stopped for several other reasons, including that a node has too few cases.

Once a terminal node is found we must decide how to classify all cases falling within it. One simple criterion is the plurality rule: the group with the greatest representation determines the class assignment. CART goes a step further: because each node has the potential for being a terminal node, a class assignment is made for every node whether it is terminal or not. The rules of class assignment can be modified from simple plurality to account for the costs of making a mistake in classification and to adjust for over- or under-sampling from certain classes. A common technique among the first generation of tree classifiers was to continue splitting nodes (growing the tree) until some goodness-of-split criterion failed to be met. When the quality of a particular split fell below a certain threshold, the tree was not grown further along that branch. When all branches from the root reached terminal nodes, the tree was considered complete. Once a maximal tree is generated, it examines smaller trees obtained by pruning away branches of the maximal tree. Once the maximal tree is grown and a set of sub-trees is derived from it, CART determines the best tree by testing for error rates or costs. With sufficient data, the simplest method is to divide the sample into learning and test sub-samples. The learning sample is used to grow an overly large tree. The test sample is then used to estimate the rate at which cases are misclassified (possibly adjusted by misclassification costs). The misclassification error rate is calculated for the largest tree and also for every sub-tree. The best sub-tree is the one with the lowest or near-lowest cost, which may be a relatively small tree. Cross validation is used if data are insufficient for a separate test sample.

In the search for patterns in databases it is essential to avoid the trap of over fitting or finding patterns that apply only to the training data. CART's embedded test disciplines ensure that the patterns found will hold up when applied to new data. Further, the testing and selection of the optimal tree are an integral part of the CART algorithm. CART handles missing values in the database by substituting surrogate splitters, which are back-up rules that closely mimic the action of primary splitting rules. The surrogate splitter contains information that is typically similar to what would be found in the primary splitter.

**Hybrid CART-MARS Model**

Figure 2 illustrates the hybrid CART-MARS model for predicting the forex values. CART and MARS could be integrated to work in a cooperative or concurrent environment. In a cooperative environment, CART plays an important role during the initialization of the prediction model. CART would go to the background after supplying some important variable information to MARS for building up the model. Thereafter MARS model works independently for further prediction. This sort of combination might be useful when not much variation is expected in the forex data.

In a concurrent mode, CART and MARS are not independent. CART continuously provides intelligent variable information to improve the MARS prediction accuracy. This combination might be helpful when the forex data is continuously changing and requires constant updating of the prediction model.

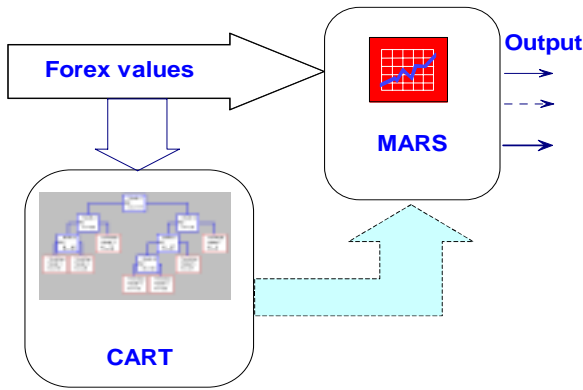

**Figure 2.** Hybrid cooperative CART-MARS model for forex monitoring

We used the cooperative model where the forex values are fed to CART to provide some additional variable information to MARS. For modeling the forex data, we supplied MARS with the "node" information generated by CART.

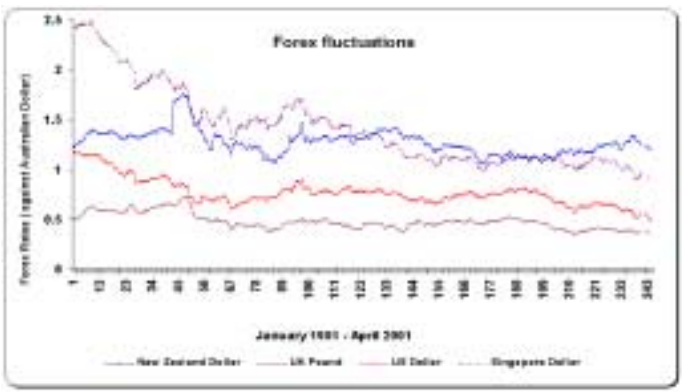

**Figure 3.** Forex values from January 1981 – April 2001 for four currencies.

## IV. EXPERIMENTATION SETUP – TRAINING AND PERFORMANCE EVALUATION

The data for our study were the monthly average forex rates from January 1981 to April 2001. We considered the exchange rates of the Australian dollar with respect to the Japanese yen, US Dollar, UK pound, Singapore dollar and New Zealand dollar. Figure 3 shows the forex fluctuations during the period January 1981 – April 2001 for the four different currencies. The experimental system consists of two stages: training the prediction systems and performance evaluation. For training the neural network, neuro-fuzzy model, MARS, CART and hybrid CART-MARS model, we choosed the "*month*" and "*previous month's forex rate*" as input variables and the "*current months forex rate*" as output variable. We randomly extracted 70% of the data for training the prediction models and the remaining for testing purposes. The test data was then passed through the trained models to evaluate the prediction efficiency. Our objective is to develop an efficient and accurate forex prediction system capable of producing a reliable forecast .The required time-resolution of the forecast is monthly, and the required time-span of the forecast is one month ahead. This means that the system should be able to predict the forex rates one month ahead based on the values of the previous month. We used a Pentium II, 450 MHz platforms for simulating the models.

**Training the different Computing Models**

- **Soft computing models**

Our preliminary experiments helped us to formulate a feedforward neural network with 1 input layer, 2 hidden layers and an output layer [6-14-14-1]. Input layer consists of 6 neurons corresponding to the input variables. The first and second hidden layers consist of 14 neurons respectively using tanh-sigmoidal activation functions. Training was terminated after 2000 epochs and we achieved a training error of 0.0251. For training the neuro-fuzzy (NF) model, we used 4 gaussian membership functions for each input variables and 16 rules were learned using the hybrid training method. Training was terminated after 30 epochs. For the NF model, we achieved training RMSE of 0.0248. The developed Takagi-Sugeno FIS is illustrated in Figure 11. While the neural network took 200 seconds for 2000 epochs training, the neuro-fuzzy model took only 35 seconds for 30 epochs training.

- **MARS**

We used 30 basis functions and to obtain the best possible prediction results (lowest RMSE), we sacrificed the speed (minimum completion time). It took almost 1 second to train the different forex prediction models.

- **CART**

We selected the minimum cost tree regardless the size of the tree. For NZ dollar prediction, the developed tree has 7 terminal nodes as shown in Figure 5. Also Figure 4 illustrates the variation of error as the numbers of nodes are increased. It took 3 seconds for developing the CART model for each prediction.

- **Hybrid MARS-CART**

In the hybrid approach the data sets were first passed through CART and the node information were generated. The training data together with the node information were supplied for training MARS.

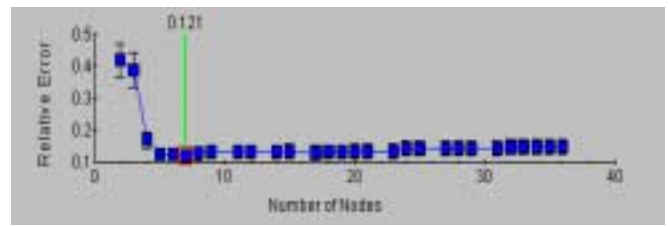

**Figure 4.** Change in relative error when the number of nodes is increased.

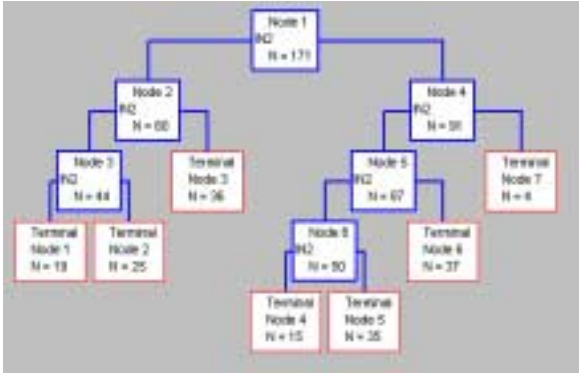

**Figure 5**. Developed regression tree for NZ dollar prediction.

## Test results

Table 1 summarizes the performances of neural network, neuro-fuzzy model, MARS, CART and hybrid CART-MARS on the test set data. Figure 6, 7, 8, 9 and 10 illustrates the test results for forex prediction using MARS and CART and hybrid CART-MARS. In Figures 6-10, the actual predicted values by each technique is plotted against the desired value for each currency

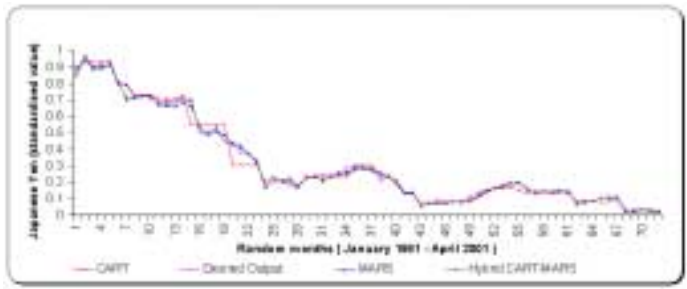

**Figure 6**. Test results for Japanese Yen

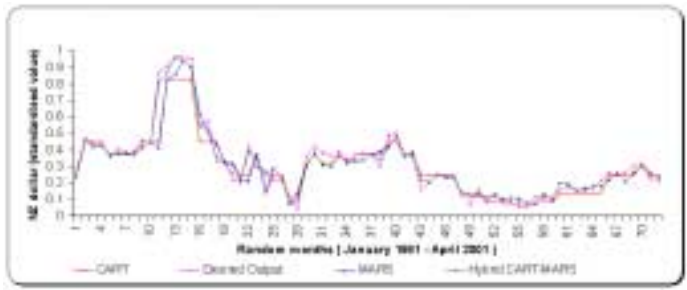

**Figure 7**. Test results for New Zealand dollar

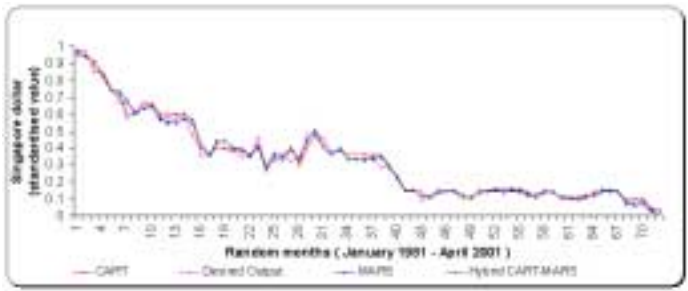

**Figure 8.** Test results for Singapore dollar

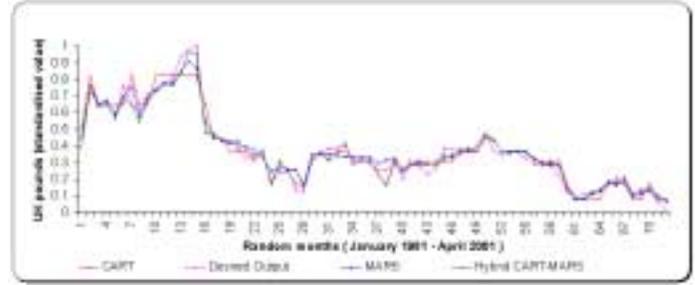

**Figure 9**. Test results for UK pounds

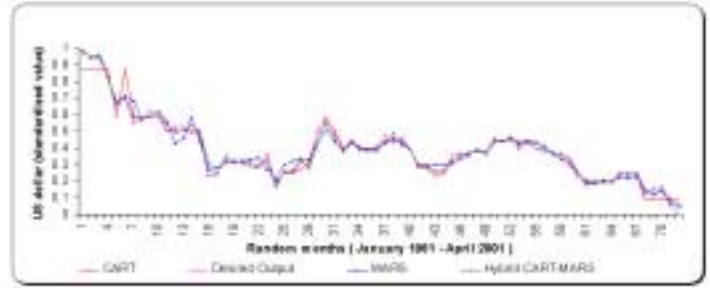

**Figure 10.** Test results for US dollar

## V. CONCLUSIONS

In this paper, we have investigated the performance of neural network, neuro-fuzzy system, MARS, CART and a hybrid CART-MARS technique for predicting the monthly average forex rates of US dollar, UK pounds, Singapore dollar, New Zealand dollar and Japanese yen with respect to Australian dollar. RMSE values of the test results reveal that, in four cases the hybrid approach (hybrid CART-MARS and NF model) performed better than the other techniques when trained/tested independently. The NZ dollar data could be best modeled by neural networks and neuro-fuzzy system. For the prediction of UK pounds, neural networks gave the lowest RMSE while for NZ dollar the NF model performed the best. It is difficult to comment on the results theoretically as very often the performance directly depends on the profile of the data itself.

The plotted figures (6-10) also graphically reveal about the quality of the prediction. While the considered soft computing models requires several iterations of training the MARS/CART hard computing approach works on a one pass training approach. Hence compared to soft computing, an important advantage of the considered hard computing approach is the speed and accuracy. Soft computing models on the other hand are very robust. Hence superiority of the soft computing models will be on the robustness in particularly the easy interpretability (*if-then* rules) of the neuro-fuzzy models as shown in Figure 11.

In the future, we also plan to investigate the hybridization of hard computing techniques with popular soft computing approaches like neural networks and neuro-fuzzy systems.

**TABLE 1. TEST RESULTS AND PERFORMANCE COMPARISON USING HYBRID SOFT AND HARD COMPUTING TECHNIQUES**

|  | Japanese Yen | US $ | UK £ | Singapore $ | New Zealand $ |
|---|---|---|---|---|---|
|  | **MARS** | | | | |
|  | 0.023 | 0.039 | 0.0478 | 0.028 | 0.049 |
|  | **CART** | | | | |
|  | 0.037 | 0.037 | 0.063 | 0.033 | 0.041 |
|  | **Hybrid CART - MARS** | | | | |
| **Test set RMSE** | 0.016 | 0.027 | 0.035 | 0.026 | 0.035 |
|  | **Artificial neural network** | | | | |
|  | 0.028 | 0.0340 | 0.023 | 0.030 | 0.021 |
|  | **Neuro-fuzzy model** | | | | |
|  | 0.026 | 0.0340 | 0.037 | 0.029 | 0.020 |

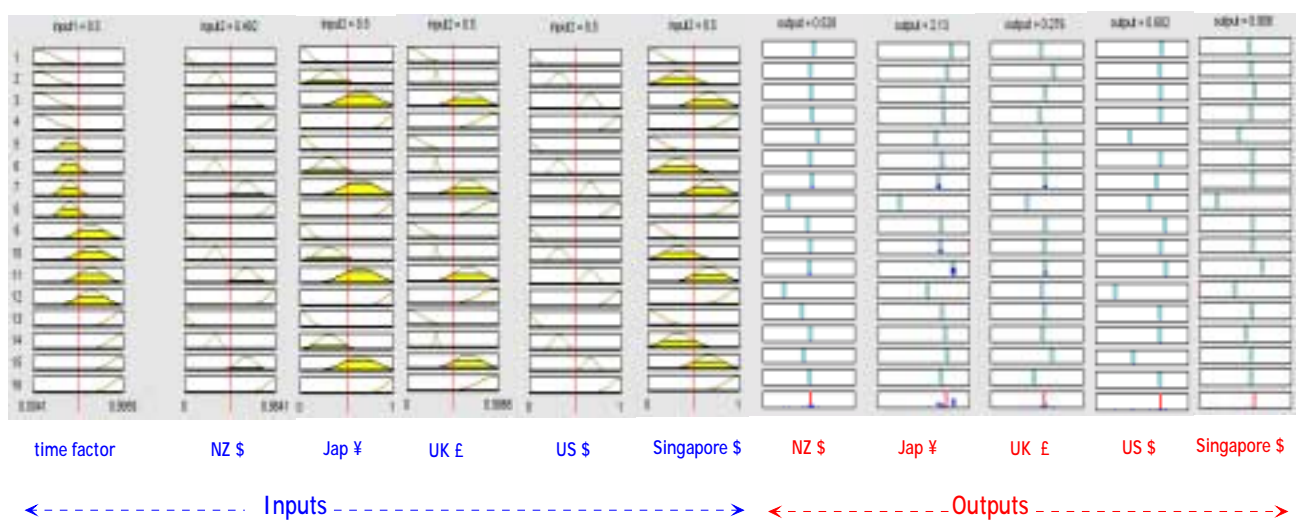

**Figure 11.** Developed Takagi-Sugeno type fuzzy inference model for forex prediction

## ACKNOWLEDGEMENTS

Author would like to thank Dr. Dan Steinberg, CEO of Salford Systems Inc. USA, for giving valuable comments on the experiments related to hybrid MARS - CART. Also thanks to Edmond Bosworth (Group Capital Management of National Australia Bank Ltd.) for the fruitful discussions and suggestions, which has provided more technical insights and sound understanding of the forex market.